\PassOptionsToPackage{unicode}{hyperref}
\PassOptionsToPackage{hyphens}{url}
\documentclass[
]{article}
\usepackage{amsmath,amssymb}
\usepackage{iftex}
\ifPDFTeX
  \usepackage[T1]{fontenc}
  \usepackage[utf8]{inputenc}
  \usepackage{textcomp} 
\else 
  \usepackage{unicode-math} 
  \defaultfontfeatures{Scale=MatchLowercase}
  \defaultfontfeatures[\rmfamily]{Ligatures=TeX,Scale=1}
\fi
\usepackage{lmodern}
\ifPDFTeX\else
\fi
\IfFileExists{upquote.sty}{\usepackage{upquote}}{}
\IfFileExists{microtype.sty}{
  \usepackage[]{microtype}
  \UseMicrotypeSet[protrusion]{basicmath} 
}{}
\makeatletter
\@ifundefined{KOMAClassName}{
  \IfFileExists{parskip.sty}{%
    \usepackage{parskip}
  }{
    \setlength{\parindent}{0pt}
    \setlength{\parskip}{6pt plus 2pt minus 1pt}}
}{
  \KOMAoptions{parskip=half}}
\makeatother
\usepackage{xcolor}
\usepackage{longtable,booktabs,array}
\usepackage{calc} 
\usepackage{etoolbox}
\makeatletter
\patchcmd\longtable{\par}{\if@noskipsec\mbox{}\fi\par}{}{}
\makeatother
\IfFileExists{footnotehyper.sty}{\usepackage{footnotehyper}}{\usepackage{footnote}}
\makesavenoteenv{longtable}
\usepackage{graphicx}
\makeatletter
\def\maxwidth{\ifdim\Gin@nat@width>\linewidth\linewidth\else\Gin@nat@width\fi}
\def\maxheight{\ifdim\Gin@nat@height>\textheight\textheight\else\Gin@nat@height\fi}
\makeatother
\setkeys{Gin}{width=\maxwidth,height=\maxheight,keepaspectratio}
\makeatletter
\def\fps@figure{htbp}
\makeatother
\providecommand{\tightlist}{%
  \setlength{\itemsep}{0pt}\setlength{\parskip}{0pt}}
\NewDocumentCommand\citeproctext{}{}

\makeatletter
 \let\@cite@ofmt\@firstofone
 \def\@biblabel#1{}
 \def\@cite#1#2{{#1\if@tempswa , #2\fi}}
\makeatother
\newlength{\cslhangindent}
\newlength{\csllabelwidth}
\newenvironment{CSLReferences}[2] 
 {\begin{list}{}{%
  \setlength{\itemindent}{0pt}
  \setlength{\leftmargin}{0pt}
  \setlength{\parsep}{0pt}
  \ifodd #1
   \setlength{\leftmargin}{\cslhangindent}
   \setlength{\itemindent}{-1\cslhangindent}
  \fi
  \setlength{\itemsep}{#2\baselineskip}}}
 {\end{list}}
\usepackage{calc}

\usepackage{amsmath}
\ifLuaTeX
  \usepackage{selnolig}  
\fi
\usepackage{bookmark}
\IfFileExists{xurl.sty}{\usepackage{xurl}}{} 
\hypersetup{
  pdftitle={Climate-Driven Doubling of U.S. Maize Loss Probability: Interactive Simulation with Neural Network Monte Carlo},
  hidelinks,
  pdfcreator={LaTeX via pandoc}}

\title{Climate-Driven Doubling of U.S. Maize Loss Probability:
Interactive Simulation with Neural Network Monte Carlo}

\usepackage{authblk}
\usepackage{orcidlink}

\author[1]{A Samuel Pottinger \orcidlink{0000-0002-0458-4985}}
\author[2]{Lawson Connor \orcidlink{0000-0001-5951-5752}}
\author[1]{Brookie Guzder-Williams \orcidlink{0000-0001-6855-8260}}
\author[1]{Maya Weltman-Fahs}
\author[1]{Nick Gondek \orcidlink{0009-0007-7431-4669}}
\author[3]{Timothy Bowles \orcidlink{0000-0002-4840-3787}}

\affil[1]{Eric and Wendy Schmidt Center for Data Science and
Environment, University of California Berkeley, Berkeley 94720, CA, USA}
\affil[2]{Department of Agricultural Economics and Agribusiness,
University of Arkansas, Fayetteville 72701, AR, USA}
\affil[3]{Department of Environmental Science, Policy \& Management,
University of California Berkeley, Berkeley 94720, CA, USA}

\date{2024-12-17}

\begin{document}
\maketitle

\textbf{Abstract:} Climate change not only threatens agricultural
producers but also strains related public agencies and financial
institutions. These important food system actors include government
entities tasked with insuring grower livelihoods and supporting response
to continued global warming. We examine future risk within the U.S. Corn
Belt geographic region for one such crucial institution: the U.S.
Federal Crop Insurance Program. Specifically, we predict the impacts of
climate-driven crop loss at a policy-salient ``risk unit'' scale. Built
through our presented neural network Monte Carlo method, simulations
anticipate both more frequent and more severe losses that would result
in a costly doubling of the annual probability of maize Yield Protection
insurance claims at mid-century. We also provide an open source pipeline
and interactive visualization tools to explore these results with
configurable statistical treatments. Altogether, we fill an important
gap in current understanding for climate adaptation by bridging existing
historic yield estimation and climate projection to predict crop loss
metrics at policy-relevant granularity.

\section{Introduction}\label{introduction}

Public institutions such as government-supported crop insurance play an
important role in agricultural stability across much of the world (Mahul
and Stutley 2010). To inform climate adaptation efforts, we add to
existing work regarding global warming impacts to these essential food
systems actors (Diffenbaugh, Davenport, and Burke 2021) by providing a
neural network Monte Carlo method which we use to examine the U.S.
Federal Crop Insurance Program inside the U.S. Corn Belt geographic
region. Building upon prior climate projections (Williams et al. 2024)
and remote sensing yield estimations (D. B. Lobell et al. 2015), these
maize loss projections enable prediction of future insurance indemnity
claims at an institutionally-relevant spatial scale.

\subsection{Background}\label{background}

Global warming threatens production of key staple crops, including maize
(Rezaei et al. 2023). Climate variability already drives a substantial
proportion of year-to-year crop yield variation (Ray et al. 2015) and
continued climate change may reduce planet-wide maize yields by up to
24\% by the end of this century (Jägermeyr et al. 2021). The growing
frequency and severity of stressful weather conditions (Dai 2013) to
which maize is increasingly susceptible (D. B. Lobell, Deines, and
Tommaso 2020) pose not only a threat to farmers' revenue (Sajid et al.
2023) but also strain the institutions established to safeguard those
producers (Hanrahan 2024). These important organizations are also often
tasked with supporting the food system through evolving growing
conditions and the impacts of climate change (RMA 2022).

Within this context, the United States of America is the world's largest
maize producer and exporter (Ates 2023). Its government-backed Federal
Crop Insurance Program covers a large share of this growing risk (Tsiboe
and Turner 2023). The costs of crop insurance in the U.S. have already
increased by 500\% since the early 2000s with annual indemnities
reaching \$19B in 2022 (Schechinger 2023). Furthermore, retrospective
analysis attributes 19\% of ``national-level crop insurance losses''
between 1991 and 2017 to climate warming, an estimate rising to 47\%
during the drought-stricken 2012 growing season (Diffenbaugh, Davenport,
and Burke 2021). Looking forward, Li et al. (2022) show progressively
higher U.S. maize loss rates as warming elevates.

\subsection{Prior work}\label{prior-work}

Modeling possible changes in frequency and severity of crop loss events
that trigger indemnity claims is an important step to prepare for the
future impacts of global warming. Related studies have predicted changes
in crop yields at broad scales such as the county-level (Leng and Hall
2020) and have estimated climate change impacts to U.S. maize within
whole-sector or whole-economy analysis (Hsiang et al. 2017). These
efforts include traditional statistical models (D. B. Lobell and Burke
2010) as well as an increasing body of work favoring machine learning
approaches (Leng and Hall 2020). Finally, the literature also consider
how practice-specific insurance subsidies intersect with grower
practices (Connor, Rejesus, and Yasar 2022; Wang, Rejesus, and Aglasan
2021; Chemeris, Liu, and Ker 2022) and observed resilience (Renwick et
al. 2021; Manski et al. 2024).

Despite these prior contributions, important programs often include
highly localized variables such as an individual farm's last ten years
of yield for a specific crop (RMA 2008). Therefore, to inform policy,
research must include more granular models than previous studies (Leng
and Hall 2020) and, in addition to predicting yield (D. B. Lobell et al.
2015; Jägermeyr et al. 2021; Ma et al. 2024), need to simulate insurance
instrument mechanics. Of particular interest, we fill a need for
climate-aware simulations of loss probability and severity within a
``risk'' or ``insured'' unit, a geographic scale referring to a set of
agricultural fields that are insured together (FCIC 2020).

\subsection{Contribution}\label{contribution}

We address this need for institutionally-relevant granular future loss
prediction through neural network Monte Carlo. We provide these
projections at the policy-relevant risk unit scale, probabilistically
forecasting institution-relevant outcome metrics under climate change.
We focus on the U.S. Corn Belt, a 9 state region within the United
States essential to the nation's maize crop (Green et al. 2018). Within
this agriculturally important area, we specifically model the Yield
Protection plan, one of the options under the popular Multi-Peril Crop
Insurance Program (RMA 2024). Furthermore, by contrasting results to a
``counterfactual'' which does not include further climate warming, we
quantitatively highlight the insurer-relevant effects of climate change.
Trained on remote sensed maize yield estimations (D. B. Lobell et al.
2015), these models project future insurance outcomes at approximately
one and three decades (Williams et al. 2024).

\section{Methods}\label{methods}

We first build predictive models of maize yield distributions using a
neural network at an insurer-relevant spatial scale before simulating
changes to yield losses under different climate conditions with Monte
Carlo. From these results, we calculate the probability and severity of
indemnity claims.

\subsection{Definitions}\label{definitions}

Before modeling these systems, we articulate mathematical definitions of
domain-specific concepts and policy instruments. First, insurers pay out
based on the magnitude of a yield loss across the aggregation of all of
the fields in an insured unit. This covered loss (\(l\)) is defined as
the difference between actual yield (\(y_{actual}\)) and a guarantee
threshold set by a coverage level (\(c\)) which is a percentage of an
expected yield (\(y_{expected}\)) (RMA 2008).

\begin{equation}l = max(c * y_{expected} - y_{actual}, 0)\label{eq:loss}\end{equation}

Note that \(y_{expected}\) is typically the average of the 10 most
recent years of yield for the insured crop (RMA 2008).

\begin{equation}y_{expected} = \frac{y_{historic}[-d:]}{d}\label{eq:expected}\end{equation}

Next, we define the probability of experiencing a loss that may incur a
Yield Protection claim (\(p_{l}\)).

\begin{equation}p_{l} = P(\frac{y_{actual} - y_{expected}}{y_{expected}} < c - 1) = P(y_{\Delta\%} < c - 1)\label{eq:ploss}\end{equation}

Generally, the severity (\(s\)) of a loss when it occurs defines the
size of the claim.

\begin{equation}s = \max(-1 * y_{\Delta\%} - (1 - c), 0)\label{eq:severity}\end{equation}

Our supplemental materials include derivations and alternatives. We
present results using the more common (FCIC 2023) 75\% coverage limit
(\(c=0.75\)) but our interactive tools (Pottinger et al. 2024b) explore
other coverage levels.

\subsection{Data}\label{data}

As Yield Protection operates at the level of a risk unit, modeling these
formulations requires highly local yield and climate information.
Therefore, we use maize yield estimates from the Scalable Crop Yield
Mapper (SCYM) approach of D. B. Lobell et al. (2015). These SCYM yield
estimations from 1999 to 2022 at 30m resolution across the US Corn Belt
are derived from remote sensing and benefit from substantial validation
efforts (Deines et al. 2021). Meanwhile, we use climate data from
CHC-CMIP6 (Williams et al. 2024) which, at daily 0.05 degree or
approximately 5km scale, offers both historic data on growing conditions
from 1983 to 2016 as well as future projections with a 2030 and a 2050
series each containing multiple years. In choosing from its two
available scenarios, we prefer the ``intermediate'' SSP245 within
CHC-CMIP6 over SSP585 per the advice of Hausfather and Peters (2020).
This offers the following daily climate variables for modeling:
precipitation, temperature (min and max), relative humidity (average,
peak), heat index, wet bulb temperature, vapor pressure deficit, and
saturation vapor pressure. Note that we prefer SCYM over recent
alternatives (Ma et al. 2024) given temporal overlap with CHC-CMIP6.

\subsubsection{Neighborhoods}\label{neighborhoods}

We align these variables to a common grid in order to create the
discrete instances needed for model training and evaluation. More
specifically, we create ``neighborhoods'' (Manski et al. 2024) of
geographically proximate fields paired with climate data through 4
character geohashing\footnote{This algorithm creates hierarchical grid
  cells where each point is assigned a unique string through hashing.
  For example, the first 4 characters identifies a grid cell (approx 28
  by 20 km) which contains all points with the same first 4 characters
  of their geohash. We evaluate alternative neighborhood sizes (number
  of geohash characters) in our interactive tools.} (Niemeyer 2008). We
simulate units within each of these cells by sampling SCYM pixels within
each neighborhood to approximate risk unit size.

\subsubsection{Yield deltas}\label{yield-deltas}

Having created these spatial groups, we model against SCYM-observed
deviations from yield expectations
(\(\frac{y_{actual} - y_{expected}}{y_{expected}}\)) which can be used
to calculate loss probability (\(l\)) and severity (\(s\)). Reflecting
the mechanics of Yield Protection policies, this step converts to a
distribution of changes or ``yield deltas'' relative to the average
production histories (APH).

\subsection{Regression}\label{regression}

We next build predictive models for distributions of yield deltas.

\subsubsection{Input vector}\label{input-vector}

We predict yield delta distributions per year ahead of Monte Carlo
simulations. To predict this distribution, we describe each of the 9
CHC-CMIP6 variables as min, max, mean, and standard deviation of each
month's daily values. We also input year and baseline variability in the
form of neighborhood historic absolute yield mean (\(y_{\mu-historic}\))
and standard deviation (\(y_{\sigma-historic}\)). See interactive tools
(Pottinger et al. 2024b) for further exploration.

\subsubsection{Response vector}\label{response-vector}

Prior work suggests that yields follow a beta distribution (Nelson 1990)
but the expected shape of changes to yield (yield deltas) is unknown.
Therefore, our open source pipeline can predict shape
parameters\footnote{The neural network predicts 2 parameters for normal
  (mean, std) and 4 for beta (center, scale, a, b) (SciPy 2024). This
  use of summary statistics helps ensure appropriate dimensionality for
  the dataset size (Alwosheel, van Cranenburgh, and Chorus 2018).} for
either a normal distribution or beta distribution. We choose the
appropriate shape by calculating the skew and kurtosis of the observed
yield deltas distributions, using the normal distribution if meeting
approximate normality per H.-Y. Kim (2013) or beta distribution
otherwise.

\subsubsection{Neural network}\label{neural-network}

Our regressors (\(f\)) use neighborhood-level climate variables (\(C\))
and historic yield information to predict future yield changes
(\(y_{\Delta\%}\)) per year. We preprocess these inputs using z score
normalization (Y.-S. Kim et al. 2024).

\begin{equation}f(C_z, y_{\mu-historic-z}, y_{\sigma-historic-z}) \hat= y_{\Delta\%}(x) = \frac{y_{actual} - y_{expected}}{y_{expected}}\label{eq:nn}\end{equation}

Note that we use machine learning per the advice of Leng and Hall (2020)
and van Klompenburg, Kassahun, and Catal (2020). In addition to possibly
better out-of-sample estimation relative to other similar approaches
(Mwiti 2023), we specifically use feed forward artificial neural
networks (Baheti 2021) as they support multi-variable output within a
single model, predicting distribution parameters together in the same
network as opposed to some other machine learning options which must
predict them separately (Brownlee 2020b).

\begin{longtable}[]{@{}
  >{\raggedright\arraybackslash}p{(\columnwidth - 6\tabcolsep) * \real{0.1611}}
  >{\raggedright\arraybackslash}p{(\columnwidth - 6\tabcolsep) * \real{0.1216}}
  >{\raggedright\arraybackslash}p{(\columnwidth - 6\tabcolsep) * \real{0.4073}}
  >{\raggedright\arraybackslash}p{(\columnwidth - 6\tabcolsep) * \real{0.3100}}@{}}
\caption{Parameters which we try in different permutations to find an
optimal configuration.
\label{tbl:sweepparam}}\label{tbl:sweepparam}\tabularnewline
\toprule\noalign{}
\begin{minipage}[b]{\linewidth}\raggedright
\textbf{Parameter}
\end{minipage} & \begin{minipage}[b]{\linewidth}\raggedright
\textbf{Options}
\end{minipage} & \begin{minipage}[b]{\linewidth}\raggedright
\textbf{Description}
\end{minipage} & \begin{minipage}[b]{\linewidth}\raggedright
\textbf{Purpose}
\end{minipage} \\
\midrule\noalign{}
\endfirsthead
\toprule\noalign{}
\begin{minipage}[b]{\linewidth}\raggedright
\textbf{Parameter}
\end{minipage} & \begin{minipage}[b]{\linewidth}\raggedright
\textbf{Options}
\end{minipage} & \begin{minipage}[b]{\linewidth}\raggedright
\textbf{Description}
\end{minipage} & \begin{minipage}[b]{\linewidth}\raggedright
\textbf{Purpose}
\end{minipage} \\
\midrule\noalign{}
\endhead
\bottomrule\noalign{}
\endlastfoot
Layers & 1 - 6 & Number of feed forward layers to include where 2 layers
include 32 and then 8 nodes while 3 layers include 64, 32, and 8. Layer
sizes are \{512, 256, 128, 64, 32, 8\}. & More layers might allow
networks to learn more sophisticated behaviors but also might overfit to
input data. \\
Dropout & 0.00, 0.01, 0.05, 0.10, 0.50 & This dropout rate applies
across all hidden layers. & Random disabling of neurons may address
overfitting. \\
L2 & 0.00, 0.05, 0.10, 0.15, 0.20 & This L2 regularization strength
applies across all hidden layer neuron connections. & Penalizing
networks with edges that are ``very strong'' may confront overfitting
without changing the structure of the network itself. \\
Attr Drop & 9 & Retraining where the sweep individually drops each of
the input distributions or year or keeps all inputs. & Removing
attributes helps determine if an input may be unhelpful. \\
\end{longtable}

We ``grid search'' (Joseph 2018) in order to find a suitable combination
of neural network hyper-parameters, trying hundreds of
permutations\footnote{All non-output neurons use Leaky ReLU activation
  per Maas, Hannun, and Ng (2013) and we use AdamW optimizer (Kingma and
  Ba 2014; Loshchilov and Hutter 2017).} from Table \ref{tbl:sweepparam}
and selecting an ideal configuration based on performance. Finally, with
meta-parameters chosen, we retrain on all available data ahead of
simulations.

\subsection{Evaluation}\label{evaluation}

We choose our model using each candidate's ability to predict into
future years, a task representative of the Monte Carlo simulations
(Brownlee 2020a):

\begin{itemize}
\tightlist
\item
  \textbf{Training} on all data between 1999 to 2012 inclusive.
\item
  \textbf{Validation} on 2014 and 2016 to compare candidates from grid
  search.
\item
  \textbf{Test} on 2013 and 2015 which serve as a fully hidden set,
  estimating how the chosen model may perform in the future.
\end{itemize}

Having performed model selection, we further evaluate our chosen
regressor through additional tests which more practically estimate
performance in different ways one may consider using this method (see
Table \ref{tbl:posthoc}).

\begin{longtable}[]{@{}
  >{\raggedright\arraybackslash}p{(\columnwidth - 6\tabcolsep) * \real{0.2535}}
  >{\raggedright\arraybackslash}p{(\columnwidth - 6\tabcolsep) * \real{0.2007}}
  >{\raggedright\arraybackslash}p{(\columnwidth - 6\tabcolsep) * \real{0.3768}}
  >{\raggedright\arraybackslash}p{(\columnwidth - 6\tabcolsep) * \real{0.1690}}@{}}
\caption{Overview of trials after model selection.
\label{tbl:posthoc}}\label{tbl:posthoc}\tabularnewline
\toprule\noalign{}
\begin{minipage}[b]{\linewidth}\raggedright
\textbf{Trial}
\end{minipage} & \begin{minipage}[b]{\linewidth}\raggedright
\textbf{Purpose}
\end{minipage} & \begin{minipage}[b]{\linewidth}\raggedright
\textbf{Train}
\end{minipage} & \begin{minipage}[b]{\linewidth}\raggedright
\textbf{Test}
\end{minipage} \\
\midrule\noalign{}
\endfirsthead
\toprule\noalign{}
\begin{minipage}[b]{\linewidth}\raggedright
\textbf{Trial}
\end{minipage} & \begin{minipage}[b]{\linewidth}\raggedright
\textbf{Purpose}
\end{minipage} & \begin{minipage}[b]{\linewidth}\raggedright
\textbf{Train}
\end{minipage} & \begin{minipage}[b]{\linewidth}\raggedright
\textbf{Test}
\end{minipage} \\
\midrule\noalign{}
\endhead
\bottomrule\noalign{}
\endlastfoot
Random Assignment & Evaluate ability to predict generally. & Random 75\%
of year / geohash combinations such that a geohash may be in training
one year and test another. & The remaining 25\% of year / region
combinations. \\
Temporal Displacement & Evaluate ability to predict into future years. &
All data from 1999 to 2013 inclusive. & All data 2014 to 2016
inclusive. \\
Spatial Displacement & Evaluate ability to predict into unseen
geographic areas. & All 4 character geohashes in a randomly chosen 75\%
of 3 character regions. & Remaining 25\% of regions. \\
Climatic Displacement & Evaluate ability to predict into out of sample
growing conditions. & All years but 2012. & 2012 (unusually dry /
hot) \\
\end{longtable}

These post-hoc trials use only training and test sets as we fully
retrain models using unchanging sweep-chosen hyper-parameters as
described in Table \ref{tbl:sweepparam}. Note that some of these tests
use ``regions'' which we define as all geohashes sharing the same first
three characters, creating a grid of 109 x 156 km cells (Haugen 2020)
each including all neighborhoods (4 character geohashes) found within.

\subsection{Simulation}\label{simulation}

\begin{figure}
\centering
\includegraphics[width=0.8\textwidth,height=\textheight]{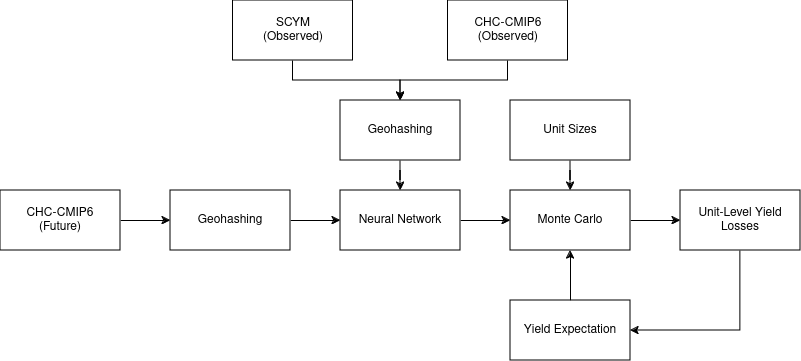}
\caption{Model pipeline overview diagram. Code released as open
source.}\label{fig:pipeline}
\end{figure}

As described in Figure \ref{fig:pipeline}, neural network predictions of
future yield delta distributions feed into Monte Carlo simulations
(Metropolis 1987; Kwiatkowski 2022) which estimate probabilities and
severity of losses at the risk unit scale. This operation happens for
each of the 17 years\footnote{CHC-CMIP6 predicts conditions for a 2030
  and a 2050 series. These predictions are provided annually as
  conditions are co-correlated within a year. However, this product
  offers our modeling a sense of conditions around those timeframes but
  does not, for example, predict 2035 specifically.} found within the
2030 and 2050 CHC-CMIP6 series (Williams et al. 2024).

Within each neighborhood, this approach simulates possible risk unit
yield deltas and allows us to consider a distribution of future
outcomes. These results then enable us to make statistical statements
about systems-wide institution-relevant metrics such as claims rate
(\(p_{l}\)).

\subsubsection{Trials}\label{trials}

Each Monte Carlo trial involves multiple sampling operations. First, we
sample climate variables and model error residuals to propagate
uncertainty (Yanai et al. 2010). Next, we draw yield multiple times to
approximate the size of a risk unit with its portfolio effects. Note
that the size but not the specific location of insured units is publicly
disclosed. Therefore, we draw the geographic size of each insured unit
randomly from historic data (RMA 2024) as part of Monte Carlo. Trials
are further described in our supplemental materials.

\subsubsection{Statistical tests}\label{statistical-tests}

Altogether, this approach simulates insured units individually per year.
Having found these outcomes as a distribution, we can then evaluate
these results probabilistically. As further described in supplemental,
we determine significance both in this paper and our interactive tools
via Bonferroni-corrected (Bonferroni 1935) Mann Whitney U (Mann and
Whitney 1947) per neighborhood.

\section{Results}\label{results}

We project climate change to roughly double loss probabilities
(\(p_{l}\)) at mid-century.

\subsection{Aggregation outcomes}\label{aggregation-outcomes}

The dataset spanning 1999 to 2016 includes a median of 83k SCYM yield
estimations per neighborhood. These field-level estimations are
represented within annual neighborhood-level yield distributions. While
yield itself is often not normally distributed, nearly all yield
\emph{delta} distributions exhibit approximate normality (H.-Y. Kim
2013). Therefore, we report model outputs assuming a normal distribution
of yield deltas. However, our supplemental materials provide further
statistics and alternative beta distribution results.

\subsection{Neural network outcomes}\label{neural-network-outcomes}

With bias towards performance in mean prediction, we select 6 hidden
layers using 0.05 dropout and 0.05 L2 from our sweep with all data
attributes included. As described in supplemental, additional layers
show diminishing returns. Table \ref{tbl:trainresults} reports mean
absolute error (MAE) in yield delta percentage points
(\(|\frac{y_{actual} - y_{expected}}{y_{expected}} - y_{\Delta\% - Predicted}|\)).
Our selected model sees 6.2\% MAE when predicting neighborhood mean
change in yield (\(y_{\Delta\%}\)) and 2.0\% when predicting
neighborhood standard deviation in our fully hidden test set after
retraining with train and validation together.

\begin{longtable}[]{@{}
  >{\raggedright\arraybackslash}p{(\columnwidth - 4\tabcolsep) * \real{0.3077}}
  >{\raggedright\arraybackslash}p{(\columnwidth - 4\tabcolsep) * \real{0.3538}}
  >{\raggedright\arraybackslash}p{(\columnwidth - 4\tabcolsep) * \real{0.3385}}@{}}
\caption{Results of model training and selection.
\label{tbl:trainresults}}\label{tbl:trainresults}\tabularnewline
\toprule\noalign{}
\begin{minipage}[b]{\linewidth}\raggedright
\textbf{Set}
\end{minipage} & \begin{minipage}[b]{\linewidth}\raggedright
\textbf{MAE for Mean Prediction}
\end{minipage} & \begin{minipage}[b]{\linewidth}\raggedright
\textbf{MAE for Std Prediction}
\end{minipage} \\
\midrule\noalign{}
\endfirsthead
\toprule\noalign{}
\begin{minipage}[b]{\linewidth}\raggedright
\textbf{Set}
\end{minipage} & \begin{minipage}[b]{\linewidth}\raggedright
\textbf{MAE for Mean Prediction}
\end{minipage} & \begin{minipage}[b]{\linewidth}\raggedright
\textbf{MAE for Std Prediction}
\end{minipage} \\
\midrule\noalign{}
\endhead
\bottomrule\noalign{}
\endlastfoot
Train & 6.1\% & 2.0\% \\
Validation & 9.4\% & 3.2\% \\
Test with retrain & 6.2\% & 2.0\% \\
\end{longtable}

In addition to Table \ref{tbl:posthocresults} which evaluates regression
performance in varied test sets, our interactive tools (Pottinger et al.
2024b) and supplemental materials offer additional performance metrics.

\begin{longtable}[]{@{}
  >{\raggedright\arraybackslash}p{(\columnwidth - 6\tabcolsep) * \real{0.2333}}
  >{\raggedright\arraybackslash}p{(\columnwidth - 6\tabcolsep) * \real{0.2444}}
  >{\raggedright\arraybackslash}p{(\columnwidth - 6\tabcolsep) * \real{0.2333}}
  >{\raggedright\arraybackslash}p{(\columnwidth - 6\tabcolsep) * \real{0.2889}}@{}}
\caption{Results of tests after model selection. Tasks have a different
number of risk units within their test set based on task definition.
\label{tbl:posthocresults}}\label{tbl:posthocresults}\tabularnewline
\toprule\noalign{}
\begin{minipage}[b]{\linewidth}\raggedright
\textbf{Task}
\end{minipage} & \begin{minipage}[b]{\linewidth}\raggedright
\textbf{Test Mean Pred MdAE}
\end{minipage} & \begin{minipage}[b]{\linewidth}\raggedright
\textbf{Test Std Pred MdAE}
\end{minipage} & \begin{minipage}[b]{\linewidth}\raggedright
\textbf{\% of Units in Test Set}
\end{minipage} \\
\midrule\noalign{}
\endfirsthead
\toprule\noalign{}
\begin{minipage}[b]{\linewidth}\raggedright
\textbf{Task}
\end{minipage} & \begin{minipage}[b]{\linewidth}\raggedright
\textbf{Test Mean Pred MdAE}
\end{minipage} & \begin{minipage}[b]{\linewidth}\raggedright
\textbf{Test Std Pred MdAE}
\end{minipage} & \begin{minipage}[b]{\linewidth}\raggedright
\textbf{\% of Units in Test Set}
\end{minipage} \\
\midrule\noalign{}
\endhead
\bottomrule\noalign{}
\endlastfoot
Random & 5.0\% & 1.6\% & 15.4\% \\
Temporal & 8.3\% & 2.1\% & 17.0\% \\
Spatial & 4.7\% & 1.7\% & 24.8\% \\
Climatic & 5.2\% & 1.9\% & 5.2\% \\
\end{longtable}

\subsection{Simulation outcomes}\label{simulation-outcomes}

After retraining on all available data using the selected configuration
from our sweep, Monte Carlo simulates risk unit yield deltas from which
we derive overall system metrics like claims rate. To capture insurance
mechanics, these trials track changes to average yields over time at the
neighborhood and approximated risk unit level. Additionally, we also
sample test set model residuals to account for error. Despite the
conservative nature of the Bonferroni correction (McDonald 2014), 95.3\%
of maize acreage in SSP245 falls within a neighborhood with significant
changes to claim probability (\(p < 0.05 / n\)) at some point during the
2050 series simulations.

\begin{figure}
\centering
\includegraphics{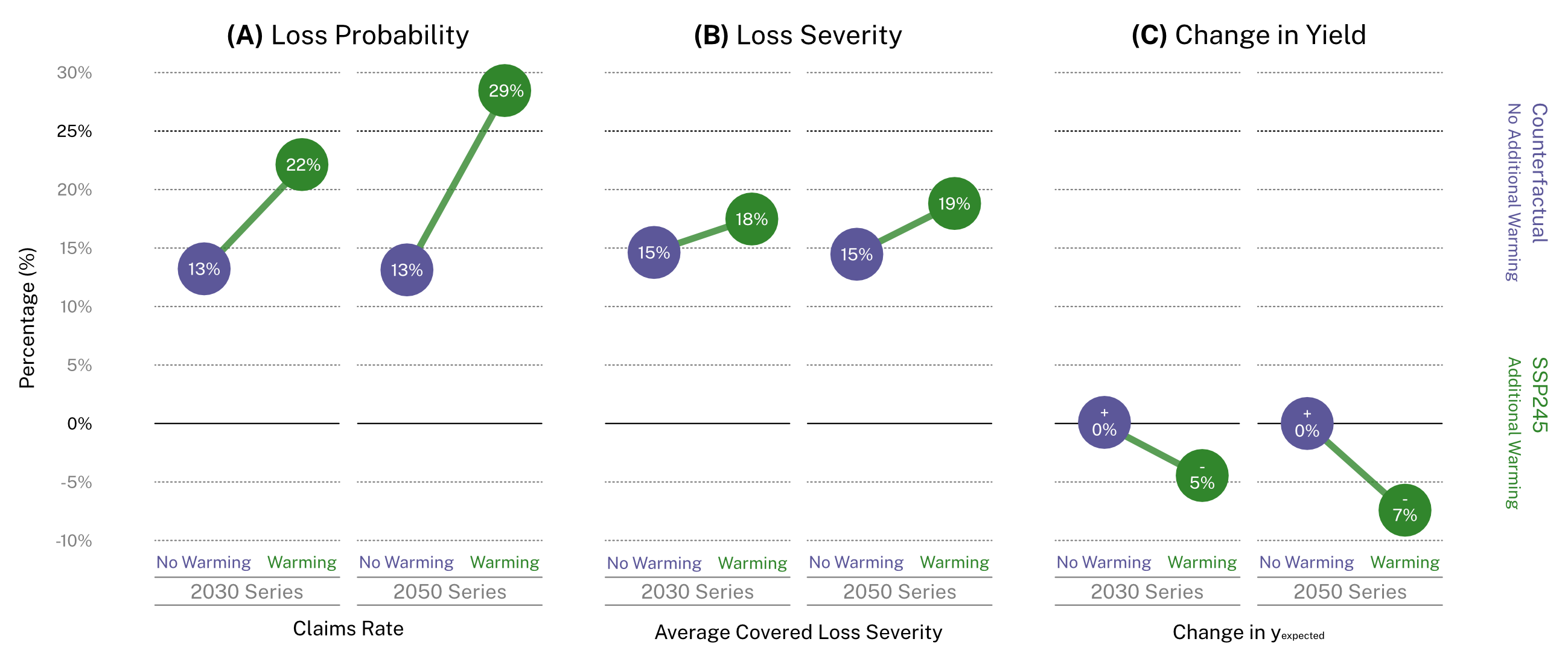}
\caption{Overview of Monte Carlo simulation results comparing SSP245
versus counterfactual for (A) loss probability, (B) loss severity, and
(C) change in average yields. Counterfactual is a future without
continued warming.}\label{fig:simresults}
\end{figure}

The claims rate elevates in the 2030 series and doubles in the 2050
timeframe when using SSP245 relative to the no further warming
counterfactual. Additionally, climate change reduces the expected
average yield and, as 2050 witnesses further warming compared to 2030,
later simulations report higher claims rates.

\section{Discussion}\label{discussion}

We observe a number of policy-relevant dynamics when simulating
insurance instrument mechanics under climate change.

\subsection{Yield expectations}\label{yield-expectations}

\begin{figure}
\centering
\includegraphics{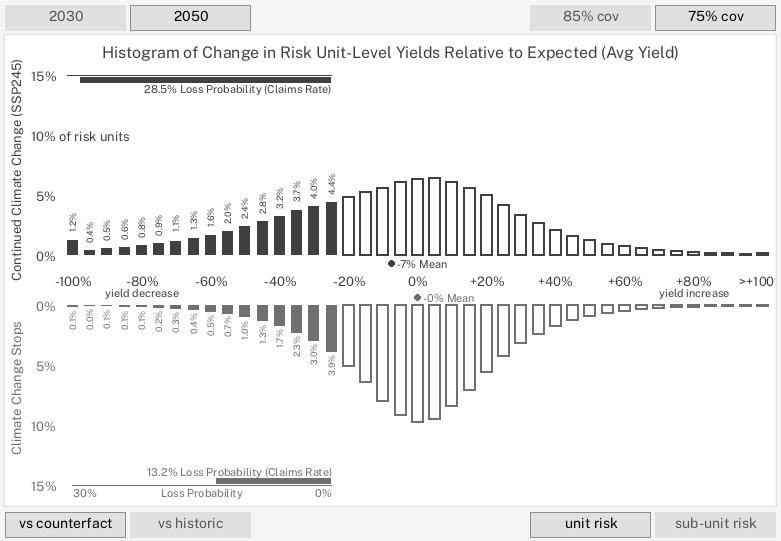}
\caption{Interactive tool screenshot showing 2050 outcomes distribution
as changes from \(y_{expected}\), plotting deltas and claims rates with
climate change on the top and without further climate change
(counterfactual) on bottom.}\label{fig:hist}
\end{figure}

Figure \ref{fig:hist} reveals possible challenges with using a simple
average in crop insurance products. While current instruments use
\(y_{expected}\) to capture changes to risk, our simulations anticipate
higher yield volatility to skew yield delta distributions such that
simulated risk units see higher claims rates despite changing
\(y_{expected}\) values.

Indeed, as further described in supplemental Table S5, 12.7\% of
neighborhoods and 9.8\% of counties under SSP245 in the 2050 series
report both increased claims rates and increased average yields. In
other words, yield volatility could allow a sharp elevation in loss
probability without necessarily decreasing overall mean yields
substantially enough to reduce claims rates through \(y_{expected}\).
These results highlight a need for future research into alternative FCIP
policy formulations, such as using historic yield variance when
establishing production histories and \(y_{expected}\).

\subsubsection{Impact to insurers}\label{impact-to-insurers}

Plans where loss is calculated against averages of historic yields may
fail to capture an increase in risk due to changing shapes of yield
delta distributions (FCIC 2020). This could allow the smoothing effect
of mean yields to mask increasing loss and insurer strain. In other
words, risk may increase at the insured unit scale in a way that is
``invisible'' to some current policy instruments.

\subsubsection{Impact to growers}\label{impact-to-growers}

Some risk mitigating practices such as regenerative agriculture trade
higher output for stability (D. Lobell et al. 2024), guarding against an
elevated probability of loss events (Renwick et al. 2021) at the cost of
a slightly reduced average (Deines et al. 2023). Therefore, our results
may indicate a mechanism for how average-based expectations could
possibly disincentivize growers from climate change preparation. That
said, we acknowledge that crop insurance effects on grower behavior
remains an area of active investigation (Connor, Rejesus, and Yasar
2022; Wang, Rejesus, and Aglasan 2021; Chemeris, Liu, and Ker 2022).

\subsection{Recent actual claims
rates}\label{recent-actual-claims-rates}

We generally predict a 13\% claims rate in 2030 and 2050
``counterfactual'' simulations which anticipate yields absent further
climate change (future conditions similar to recent past). For
comparison, the annual median of the years for which SCYM and historic
CHC-CMIP6 data are available has an actual claims rate of 14\% (RMA
2024) amid growing conditions similar to counterfactual trials.

\subsubsection{Under-estimation}\label{under-estimation}

Despite this similarity between predictions and the comparable recent
actuals, a number of difficult to model factors likely lead us to
underestimate the actual claims rate in practice. First, field-level
yield data and the actual geographically specific risk unit structure
are not currently public. Therefore, while we sample units randomly
based on expected size, growers likely optimize their own unit structure
when purchasing policies to optimize financial upside. Similarly, we do
not have the geographically specific data required to model trend
adjustment and yield exclusion options\footnote{Under certain
  conditions, trend adjustment increases \(y_{expected}\) beyond the
  historic average (Plastina and Edwards 2014) to anticipate expected
  yield improvements while exclusions remove poor years from
  \(y_{expected}\) (Schnitkey, Sherrick, and Coppess 2015).}. These
factors likely increase the actual claims rates by raising
\(y_{expected}\). See supplemental for more details.

\subsubsection{Variation}\label{variation}

While these model limitations likely overall lead to a suppression of
loss rates in our simulations relative to actuals, note that these
adjustments change over time and could cause further fluctuations
alongside growing condition variability. For example, 2014 saw a number
of statutory changes to yield exclusions (ERS 2024). In total, we
anticipate that the future will likely see substantial annual variation
similar to the recent past even as our results still capture overall
long term trends.

\subsection{Geographic bias}\label{geographic-bias}

\begin{figure}
\centering
\includegraphics{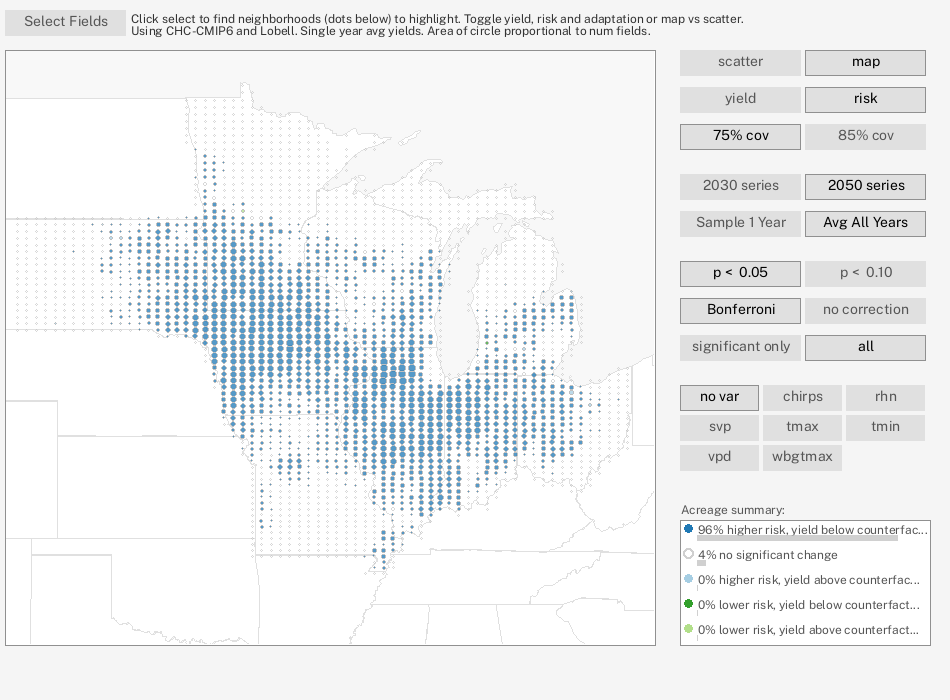}
\caption{Interactive geographic view considers different parameters and
alternative statistical treatments. Color describes type of change.
Larger dots are larger areas of maize growing activity. Graphic reveals
a possible geographic bias, especially in Iowa, Illinois, and
Indiana.}\label{fig:geo}
\end{figure}

Neighborhoods with significant results (\(p < 0.05 / n\)) may be more
common in some areas as shown in Figure \ref{fig:geo}. This spatial
pattern may partially reflect that a number of neighborhoods have less
land dedicated to maize so simulations have smaller sample sizes and
fail to reach significance. However, this may also reflect geographical
bias in altered growing conditions.

Reflecting empirical studies that document the negative impacts of heat
stress and water deficits on maize yields (Sinsawat et al. 2004; Marouf
et al. 2013), we note that spatial distribution of anticipated combined
warmer and drier conditions partially mirror areas of lower yield
predictions, possibly highlighting analogous stresses to 2012 and its
historically poor maize production (ERS 2013).

\subsection{Other limitations and future
work}\label{other-limitations-and-future-work}

We next highlight opportunities for future work.

\subsubsection{Future data}\label{future-data}

We acknowledge limitations of our findings due to constraints of the
currently available public datasets. First, though our interactive tools
consider different spatial aggregations such as 5 character (approx 4 x
5 km) geohashes, future work may consider modeling with actual reported
field-level yield data and the actual risk unit structure if later made
public. Additionally, we highlight that we focus on systematic changes
in growing conditions impacting claims rates across a broad geographic
scale. This excludes highly localized effects like certain inclement
weather which may require more granular climate predictions. This
possible future work may be relevant to programs with smaller geographic
portfolios. Next, as further described in supplemental, our model shows
signs that it is data constrained. In particular, additional years of
training data may improve performance. Our data pipeline should and can
be re-run as future versions of CHC-CMIP6 and SCYM or similar are
released. Furthermore, we also recognize that the CHC-CMIP6 2030 and
2050 series make predictions for general timeframes and not individual
specific years, a task possibly valuable for future research. Finally,
though supplemental offers further error analysis, we acknowledge that
some sources of uncertainty like from input data (SCYM and CHC-CMIP6)
cannot be quantified given currently available information.

\subsubsection{Other programs}\label{other-programs}

Outside of Yield Protection, future study could extend to the highly
related Revenue Protection form of insurance. Indeed, the yield stresses
that we describe in this model may also impact this other plan. On that
note, we include historic yield as inputs into our neural network,
allowing those data to ``embed'' adaptability measures (Hsiang et al.
2017) such as grower practices where, for example, some practices may
reduce loss events or variability (Renwick et al. 2021). That said, we
highlight that later studies looking at revenue may require additional
economic information to serve a similar role.

\subsubsection{Future benchmarking}\label{future-benchmarking}

We offer a unique focus on broad geographic institutionally-relevant
loss probability prediction at risk unit scale given remote sensed yield
estimations. Lacking a compatible study for direct contrasting of
performance measures, we invite further research on alternative
regression and simulation approaches for similar modeling objectives.
While not directly comparable, we note that D. B. Lobell and Burke
(2010) as well as Leng and Hall (2020) possibly offer precedent.

\subsection{Visualizations and
software}\label{visualizations-and-software}

In order to explore these simulations, we offer interactive open source
web-based visualizations built alongside our experiments. These both aid
us in constructing our own conclusions and allow readers to consider
possibilities and analysis beyond our own narrative. This software runs
within a web browser and is made publicly available at
https://ag-adaptation-study.org. It includes the ability to explore
alternative statistical treatments and regressor configurations as well
as generate additional geographic visualizations. Finally, in addition
to visualizations, we also offer our work as an open source data science
pipeline. This software may help aid future research into other crops
such as soy, geographic areas such as other parts of the United States
of America, other programs such as Revenue Protection, and extension of
our results as datasets are updated.

\section{Conclusion}\label{conclusion}

We present Monte Carlo on top of neural network-based regressors for
prediction of institution-relevant crop yield changes. We specifically
simulate climate-driven system-wide impacts to maize growing conditions
at a policy-relevant scale of granularity. Our results find that maize
Yield Protection claim rates may double for the U.S. Federal Crop
Insurance Program (Multi-Peril Crop Insurance) within the U.S. Corn Belt
relative to a no further warming counterfactual.

In addition to publishing our raw model outputs under a creative commons
license, we explore the specific shape of these results from the
perspective of insurance structures. First, we describe a possible
agriculturally-relevant geographic bias in climate impacts. Second, we
also highlight potential mathematical properties of interest including
increasing volatility without fully offsetting average-based yield
expectation measures. These particular kinds of changes may pose
specific threats to the current structure of existing insurance
instruments.

Altogether, this study considers how this machine learning and
interactive data science approach may understand existing food system
policy structures in the context of climate projections. Towards that
end, we release our software under permissive open source licenses and
make interactive tools available publicly at
https://ag-adaptation-study.org to further interrogate these results.
These visualizations also allow readers to explore alternatives to key
analysis parameters. This work may inform agriculture policy response to
continued climate change.

\section{Acknowledgements}\label{acknowledgements}

Study funded by the Eric and Wendy Schmidt Center for Data Science and
Environment at the University of California, Berkeley. We have no
conflicts of interest to disclose. Yield estimation data from D. B.
Lobell et al. (2015) and Deines et al. (2021) with our thanks to David
Lobell for permission. We also wish to thank Carl Boettiger, Magali de
Bruyn, Jiajie Kong, Kevin Koy, and Ciera Martinez for conversation
regarding these results. Thanks to Color Brewer (Brewer et al. 2013) and
Public Sans (General Services Administration 2024).

\textbf{Data availability statement}: Our software and pipeline source
code (Pottinger et al. 2024b) as well as our model training data and
simulation outputs (Pottinger et al. 2024a) are available on Zenodo as
open source / creative common licensed resources. Public hosted version
at https://ag-adaptation-study.org.

\section*{Works Cited}\label{works-cited}
\addcontentsline{toc}{section}{Works Cited}

\phantomsection\label{refs}

\end{document}


\maketitle

\textbf{Overview}: These supplementary materials complement
``Climate-Driven Doubling of U.S. Maize Loss Probability: Interactive
Simulation through Neural Network Monte Carlo'' to further describe the
work including statistical tests employed, simulation specfics, and the
interactive tools available at https://ag-adaptation-study.pub.

\section{Methods and data}\label{methods-and-data}

These materials start with further explanation of the methods and data
employed.

\subsection{Statistical tests}\label{statistical-tests}

To determine significance of changes to loss probability at
neighborhood-level, we use Mann Whitney U (Mann and Whitney 1947) as
variance is observed to differ between the two expected and
counterfactual sets (McDonald 2014). Given that our neural network
attempts to predict the distribution of yield deltas, we note that the
granularity of the response variable specifically may influence
statistical power for the purposes of these tests. To that end, we
observe that SYCM (Lobell et al. 2015) uses Daymet variables at 1 km
resolution (Thornton et al. 2014). Therefore, due to potential
correlation within those 1km cells, we more conservatively assume 1km
resolution to avoid artificially increasing the number of ``true'' SCYM
yield estimations per neighborhood. Finally, we recognize that we are
engaging in one statistical test per neighborhood per series (2030,
2050). We control for this through Bonferroni-correction (Bonferroni
1935).

\subsection{Insured risk unit data}\label{insured-risk-unit-data}

\begin{figure}
\centering
\includegraphics[width=0.95\textwidth,height=\textheight]{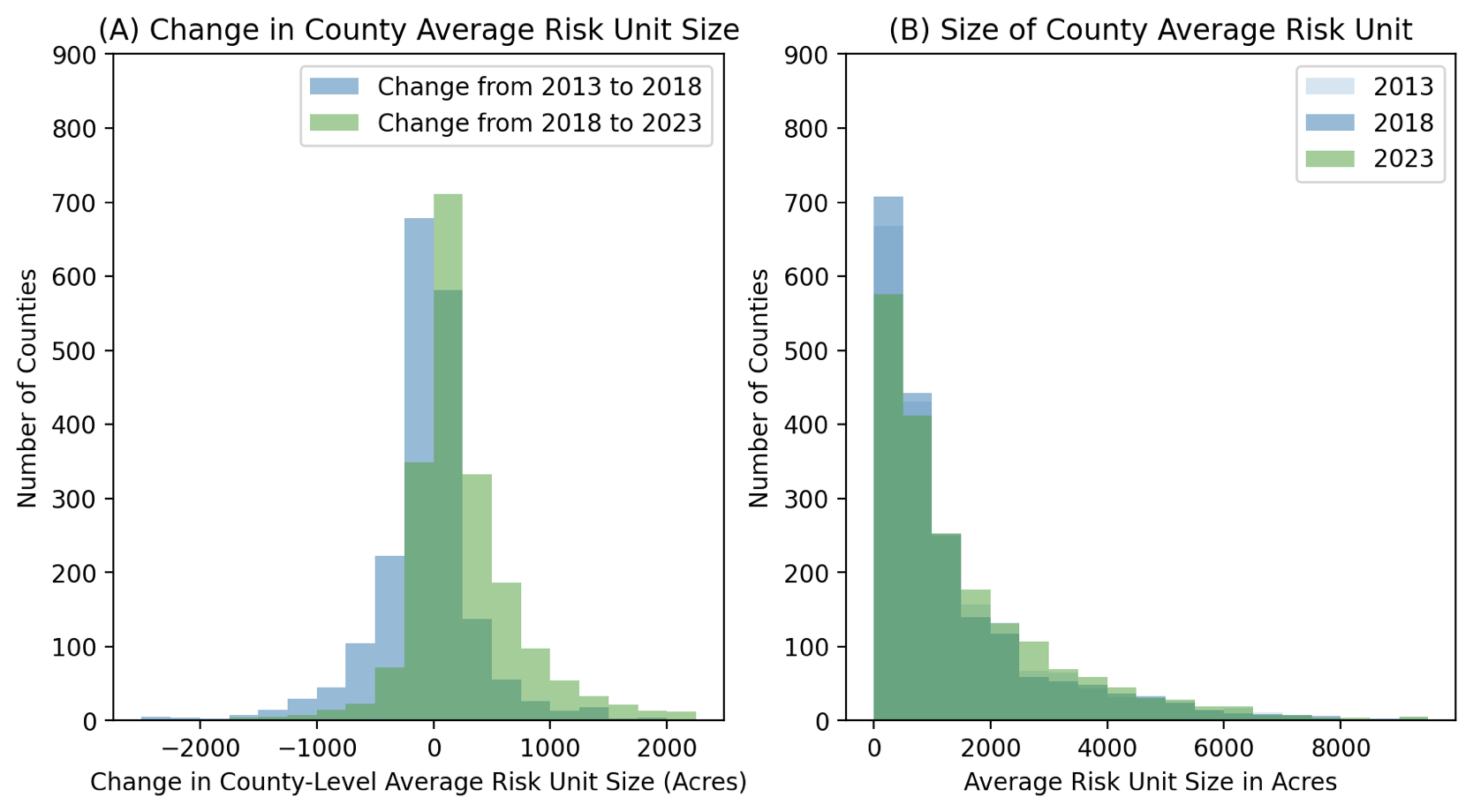}
\caption{Examination of risk unit size in years 2013, 2018, and 2023.
First, this figure shows how risk unit size changed between each year
examined (A) to highlight that the structures do evolve substantially
between years. However, these results also indicate that the overall
distribution of risk unit sizes is relatively stable (B) when considered
system-wide. Some extreme outliers not shown to preseve
detail.}\label{fig:riskunit}
\end{figure}

As visualized in the histogram displayed in Figure \ref{fig:riskunit},
the USDA provides anonymized information about risk structure (RMA
2024).

Though these data lack precise geographic specificity, the USDA
indicates the county in which these units are located. Even so, we
notice year to year instability at the county level in unit size. This
may reflect growers reconfiguring their risk structure to optimize rates
as yield profiles change over time. Altogether, this may complicate
prediction of the geographic location of larger units.

All this in mind, sampling the risk unit size at the county level likely
represents over-confidence (overfitting) to previous configurations.
Instead, we observe that the system-wide risk unit size distribution
remains relatively stable. This may suggest that, even as more local
changes to risk unit structure may be more substantial between years,
overall expectations for the size of risk units are less fluid.
Therefore, we use that larger system-wide distribution to sample risk
unit sizes within our Monte Carlo simulation instead of the county-level
distributions. This also has the effect of propogating risk unit size
uncertainty into results through the mechanics of Monte Carlo.

\subsection{Yield distributions}\label{yield-distributions}

Our treatment of yield data considers two practical constraints:

\begin{itemize}
\tightlist
\item
  Due to the size of the input dataset and engineering limitations, we
  cannot take all SCYM data per neighborhood into Monte Carlo.
\item
  We must avoid dramatic expansions to the output vector size as this
  could cause the input dataset requirements to exceed feasibility
  (Alwosheel, van Cranenburgh, and Chorus 2018).
\end{itemize}

These concerns in mind, we sample annual SCYM yields to generate yield
delta distributions which allows us to wait until the later parts of our
pipeline to make shape assumptions (normal or beta) for neighborhood or
unit-level variables. This ensures ``just in time'' that our neural
network can predict a smaller number of distribution shape parameters
while maintaing underlying shape information for as long as possible.

\subsubsection{Yield delta
distributions}\label{yield-delta-distributions}

In generating the historic yield delta distributions ahead of training
neural networks, we sample 1000 yield values per neighborhood per year
to represent a growing season\footnote{The resulting historic yield
  delta distributions are further sampled based on simulated risk unit
  size, either from historic actuals or neural network predicted
  distributions. Note that we also sample to represent historic averages
  as aggregation to \(y_{expected}\) can be subject to ``small samples''
  stochastic effects per risk unit.}. Altogether, this design avoids
needing to make distributional assumptions about yield ahead of neural
network operation while maintaining the original distributional shape.

\subsubsection{Pipline flexibility}\label{pipline-flexibility}

Our neural network requires a distributional shape assumption to
maintain a smaller output vector size. We decide the shape to predict
based on observed skew and kurtosis of yield deltas. To that end, our
open source pipeline can be run with beta or normal distribution
assumptions. The former has precedent in the literature (Nelson 1990).

\subsubsection{Practical yield delta
shape}\label{practical-yield-delta-shape}

Despite pipeline flexibility, we observe that nearly all\footnote{97\%
  of neighborhoods and maize growing acreage are approximately normal
  per Kim (2013).} yield delta distributions exhibit approximate
normality in practice (Kim 2013). Separately, as shown in Table
\ref{tbl:betadist}, using beta distributions in our neural networks
results in similar median absolute errors but elevated mean absolute
errors.

\begin{longtable}[]{@{}
  >{\raggedright\arraybackslash}p{(\columnwidth - 4\tabcolsep) * \real{0.2941}}
  >{\raggedright\arraybackslash}p{(\columnwidth - 4\tabcolsep) * \real{0.3382}}
  >{\raggedright\arraybackslash}p{(\columnwidth - 4\tabcolsep) * \real{0.3676}}@{}}
\caption{Test set performance after retraining for predicting
distribution location (mean or center) for both a normal distribution
and beta distribution assumption.
\label{tbl:betadist}}\label{tbl:betadist}\tabularnewline
\toprule\noalign{}
\begin{minipage}[b]{\linewidth}\raggedright
\textbf{Shape}
\end{minipage} & \begin{minipage}[b]{\linewidth}\raggedright
\textbf{Mean Absolute Error}
\end{minipage} & \begin{minipage}[b]{\linewidth}\raggedright
\textbf{Median Absolute Error}
\end{minipage} \\
\midrule\noalign{}
\endfirsthead
\toprule\noalign{}
\begin{minipage}[b]{\linewidth}\raggedright
\textbf{Shape}
\end{minipage} & \begin{minipage}[b]{\linewidth}\raggedright
\textbf{Mean Absolute Error}
\end{minipage} & \begin{minipage}[b]{\linewidth}\raggedright
\textbf{Median Absolute Error}
\end{minipage} \\
\midrule\noalign{}
\endhead
\bottomrule\noalign{}
\endlastfoot
Normal & 6.2\% & 5.9\% \\
Beta & 16.9\% & 7.1\% \\
\end{longtable}

Further investigation finds that that a minority population of
neighborhoods causes this swing where small changes in beta distribution
parameters can infrequently cause large error. Therefore, as prediction
of that population shows stronger performance under a normality
assumption for yield deltas, we prefer this approach in our main text.

\subsection{Neural network
configuration}\label{neural-network-configuration}

We offer additional information about the specific neural network
configuration chosen.

\subsubsection{Input vector}\label{input-vector}

Empirically leading to generally better performance, we allow the model
to use the count of growing condition estimations. This may serve as a
possible measure of uncertainty. We also allow inclusion of the year.
However, as can be executed in our open source pipeline, we find that
including absolute year generally increases overfitting. Therefore, we
use a relative measure (years since the start of the series within the
simulations). Our simulations run for 17 relative years for each series.

\subsubsection{Included years and areas}\label{included-years-and-areas}

To further document how we structure our consideration of timeseries
variables, we emphasize that we sample for 17 individual years in the
2030 CHC-CMIP6 series and 17 individual years in 2050 CHC-CMIP6 series.
Importantly, projections in these series are not necessarily intended as
specific predictions in specific years. We do not provide a year by year
timeseries for this reason. Instead, our analysis produces distributions
of anticipated outcomes at the 2030 and 2050 timeframes. Note that our
choice to create these two series follows a similar structure to
CHC-CMIP6. Finally, note that many growers engage in even simple crop
rotations so the effective average crop yield for a field used to define
yield expectations may span 10 crop years but possibly more than 10
consecutive calendar years. This is reflected in Monte Carlo sampling.

\subsubsection{Instance weight}\label{instance-weight}

We document that we build our model with instance weighting.
Specifically, we use the number (not value) of SCYM pixels in a
neighborhood to weight each neighborhood. In other words, the weight is
higher in neighborhoods with more maize growing acreage.

\subsubsection{Error and residuals}\label{error-and-residuals}

Table \ref{tbl:retrain} provides mean absolute error for the selected
model from the sweep. A drop in error observed from validation to test
with retrain\footnote{Test with retrain specifically refers to
  retraining a model from scratch using the model configuration selected
  from our hyper-parameter sweep. This training spans across both
  training and validation data together. In both the ``with retrain''
  and ``without retrain'' cases, the test set remains fully hidden.}
performance may be explained by the increased training set size. This
may indicate that the model is specifically data constrained by the
number of years available for training. Our open source data pipeline
can and will be used to rerun analysis as input datasets are updated to
include additional years in the future.

\begin{longtable}[]{@{}
  >{\raggedright\arraybackslash}p{(\columnwidth - 4\tabcolsep) * \real{0.3077}}
  >{\raggedright\arraybackslash}p{(\columnwidth - 4\tabcolsep) * \real{0.3538}}
  >{\raggedright\arraybackslash}p{(\columnwidth - 4\tabcolsep) * \real{0.3385}}@{}}
\caption{Residuals for the main training task with and without
retraining. \label{tbl:retrain}}\label{tbl:retrain}\tabularnewline
\toprule\noalign{}
\begin{minipage}[b]{\linewidth}\raggedright
\textbf{Set}
\end{minipage} & \begin{minipage}[b]{\linewidth}\raggedright
\textbf{MAE for Mean Prediction}
\end{minipage} & \begin{minipage}[b]{\linewidth}\raggedright
\textbf{MAE for Std Prediction}
\end{minipage} \\
\midrule\noalign{}
\endfirsthead
\toprule\noalign{}
\begin{minipage}[b]{\linewidth}\raggedright
\textbf{Set}
\end{minipage} & \begin{minipage}[b]{\linewidth}\raggedright
\textbf{MAE for Mean Prediction}
\end{minipage} & \begin{minipage}[b]{\linewidth}\raggedright
\textbf{MAE for Std Prediction}
\end{minipage} \\
\midrule\noalign{}
\endhead
\bottomrule\noalign{}
\endlastfoot
Train & 6.1\% & 2.0\% \\
Validation & 9.4\% & 3.2\% \\
Test with retrain & 6.2\% & 2.0\% \\
Test without retrain & 11.1\% & 2.4\% \\
\end{longtable}

The test set residuals are sampled during Monte Carlo to propogate
uncertainty. That said, we observe that a relatively small
sub-population of large percentage changes may skew results, causing the
mean and median error to diverge as shown in post-hoc tasks in Table
\ref{tbl:posthocresults}.

\begin{longtable}[]{@{}
  >{\raggedright\arraybackslash}p{(\columnwidth - 8\tabcolsep) * \real{0.1981}}
  >{\raggedright\arraybackslash}p{(\columnwidth - 8\tabcolsep) * \real{0.2075}}
  >{\raggedright\arraybackslash}p{(\columnwidth - 8\tabcolsep) * \real{0.1981}}
  >{\raggedright\arraybackslash}p{(\columnwidth - 8\tabcolsep) * \real{0.1981}}
  >{\raggedright\arraybackslash}p{(\columnwidth - 8\tabcolsep) * \real{0.1981}}@{}}
\caption{Results of tests after model selection.
\label{tbl:posthocresults}}\label{tbl:posthocresults}\tabularnewline
\toprule\noalign{}
\begin{minipage}[b]{\linewidth}\raggedright
\textbf{Task}
\end{minipage} & \begin{minipage}[b]{\linewidth}\raggedright
\textbf{Test Mean Pred MAE}
\end{minipage} & \begin{minipage}[b]{\linewidth}\raggedright
\textbf{Test Std Pred MAE}
\end{minipage} & \begin{minipage}[b]{\linewidth}\raggedright
\textbf{Test Mean Pred MdAE}
\end{minipage} & \begin{minipage}[b]{\linewidth}\raggedright
\textbf{Test Std Pred MdAE}
\end{minipage} \\
\midrule\noalign{}
\endfirsthead
\toprule\noalign{}
\begin{minipage}[b]{\linewidth}\raggedright
\textbf{Task}
\end{minipage} & \begin{minipage}[b]{\linewidth}\raggedright
\textbf{Test Mean Pred MAE}
\end{minipage} & \begin{minipage}[b]{\linewidth}\raggedright
\textbf{Test Std Pred MAE}
\end{minipage} & \begin{minipage}[b]{\linewidth}\raggedright
\textbf{Test Mean Pred MdAE}
\end{minipage} & \begin{minipage}[b]{\linewidth}\raggedright
\textbf{Test Std Pred MdAE}
\end{minipage} \\
\midrule\noalign{}
\endhead
\bottomrule\noalign{}
\endlastfoot
Random & 5.0\% & 1.6\% & 5.1\% & 1.7\% \\
Temporal & 8.3\% & 2.1\% & 7.2\% & 2.2\% \\
Spatial & 4.7\% & 1.7\% & 5.0\% & 1.7\% \\
Climatic & 5.2\% & 1.9\% & 5.2\% & 1.8\% \\
\end{longtable}

Even so, the overall error remains acceptable. In general, increased
model size is showing diminishing returns and we do not currently
consider additional layers (4 vs 5 neural network layers changes mean
prediction MAE by less than one point). Our final chosen model has the
following layer sizes: 512 neurons, 256 neurons, 128 neurons, 64
neurons, 32 neurons, 8 neurons.

\subsection{Grower behaviors}\label{grower-behaviors}

We further document some grower behaviors which may be difficult to
capture within our curent modeling structure.

\subsubsection{Historic yield averages}\label{historic-yield-averages}

Our simulations expect yield expectations to change over time. In
practice, we sample ten years of historic yields per neighborhood per
year per trial and we offset the yield deltas produced by the neural
network accordingly as the simulated timeseries progresses. This allows
for some accounting of uncertainty in yield baselines. In practice, this
means that predictions for 2030 claims rate samples the 2010 (historic)
series and 2050 samples the 2030 series. To prevent discontiniuity in
the data due to some unknown systematic model bias, the 2010 deltas are
retroactively predicted. Model error residuals are sampled in each case.

\subsubsection{Yield history
adjustments}\label{yield-history-adjustments}

In practice, the values used to set yield expectations depend on trend
adjustment (Plastina and Edwards 2014) and yield exclusions (Schnitkey,
Sherrick, and Coppess 2015) which, due to insufficient data, are left
for future work. Again, by increasing \(y_{expected}\), these may lead
to an artifical supression of our predicted claims rates.

\subsubsection{Crop rotations}\label{crop-rotations}

A large share of growers will engage in at least simple crop rotations
(Manski et al. 2024) which is important for our simulations because it
may change the locations in which maize is grown. We use SCYM to
implicitly handle this complexity. That in mind, these reported sample
sizes impact the sampling behavior during Monte Carlo and, while this
approach does not require explicit consideration of crop rotations, the
set of geohashes present in results may vary from one year to the next
in part due to this behavior.

All that said, historic locations of growth and crop rotation behavior
from the past are sampled in the future simulations. In addition to this
spatial complexity, we highlight that crop rotations mean that the last
10 years of yield data for a crop may not correspond to the last 10
calendar years. Even so, due to the ``year series'' approach in this
model, this probably has limited effect on our multi-year claims rates
estimations given estimated crop rotational complexity (Manski et al.
2024).

\subsubsection{Yield improvements}\label{yield-improvements}

While our model does not explicitly consider trend adjustment,
historically-consistent expected increases in yields outside our model
likely negate that trend adjustment. In other words, \(y_{expected}\)
under trend adjustment accounts for ``expected'' yield improvements and
may offset claims rates reductions that otherwise would be caused by
yield improvements if trend adjustment was not available. Even so,
specific investigation of this phenomenon is left for future work.

\subsubsection{Coverage levels}\label{coverage-levels}

We observe that there may be geographic bias in coverage levels. This
may include some areas with different policy availability, possibly
including geographically-biased supplemental policy usage. This results
both from grower and institutional behavior and may prove important in
specific prediction of future claims. However, lacking public data on
coverage levels chosen with geographic specificity, we respond to this
limitation by allowing for investigation of different coverage levels
within our interactive tool. Though we do not believe this to impact our
predictions of general claims probability and severity changes, this
aspect may impact research making specific annual predictions.
Therefore, we encourage future work on further investigation of coverage
level selection and its intersection with climate change.

\section{Detailed simulation results}\label{detailed-simulation-results}

For reference, we provide further detailed simulated results in Table
\ref{tbl:simresults}.

\begin{longtable}[]{@{}
  >{\raggedright\arraybackslash}p{(\columnwidth - 8\tabcolsep) * \real{0.2240}}
  >{\raggedright\arraybackslash}p{(\columnwidth - 8\tabcolsep) * \real{0.0640}}
  >{\raggedright\arraybackslash}p{(\columnwidth - 8\tabcolsep) * \real{0.2080}}
  >{\raggedright\arraybackslash}p{(\columnwidth - 8\tabcolsep) * \real{0.2640}}
  >{\raggedright\arraybackslash}p{(\columnwidth - 8\tabcolsep) * \real{0.2400}}@{}}
\caption{Details of Monte Carlo simulation results. Counterfactual is a
future without continued warming in contrast to SSP245.
\label{tbl:simresults}}\label{tbl:simresults}\tabularnewline
\toprule\noalign{}
\begin{minipage}[b]{\linewidth}\raggedright
\textbf{Scenario}
\end{minipage} & \begin{minipage}[b]{\linewidth}\raggedright
\textbf{Series}
\end{minipage} & \begin{minipage}[b]{\linewidth}\raggedright
\textbf{Unit mean yield change}
\end{minipage} & \begin{minipage}[b]{\linewidth}\raggedright
\textbf{Unit loss probability}
\end{minipage} & \begin{minipage}[b]{\linewidth}\raggedright
\textbf{Avg covered loss severity}
\end{minipage} \\
\midrule\noalign{}
\endfirsthead
\toprule\noalign{}
\begin{minipage}[b]{\linewidth}\raggedright
\textbf{Scenario}
\end{minipage} & \begin{minipage}[b]{\linewidth}\raggedright
\textbf{Series}
\end{minipage} & \begin{minipage}[b]{\linewidth}\raggedright
\textbf{Unit mean yield change}
\end{minipage} & \begin{minipage}[b]{\linewidth}\raggedright
\textbf{Unit loss probability}
\end{minipage} & \begin{minipage}[b]{\linewidth}\raggedright
\textbf{Avg covered loss severity}
\end{minipage} \\
\midrule\noalign{}
\endhead
\bottomrule\noalign{}
\endlastfoot
Historic & 2010 & 18.6\% & 7.3\% & 13.8\% \\
Counterfactual & 2030 & 0.0\% & 13.3\% & 14.7\% \\
SSP245 & 2030 & -4.5\% & 22.3\% & 17.5\% \\
Counterfactual & 2050 & -0.0\% & 13.2\% & 14.5\% \\
SSP245 & 2050 & -7.4\% & 28.5\% & 18.9\% \\
& & \(y_{\Delta \mu}\) & \(p_{l-\mu}\) & \(s_{\mu}\) \\
\end{longtable}

These results are also made available in Zenodo (A. Pottinger et al.
2024).

\subsection{Series labels}\label{series-labels}

Note that the ``2010 series'' label is used internally in our model for
consistency with 2030 and 2050 from CHC-CMIP6 though that ``2010''
language does not explicitly appear in their data model.

\subsection{Confidence}\label{confidence}

We re-execute simulations 100 times to understand variability for
system-wide metrics in Table \ref{tbl:simresults}. The range of all
standard deviations of each metric's distribution is under 0.1\% and the
range under 1\%. These tight intervals likely reflect the high degree of
aggregation represented in our system-wide metrics. However, lacking
confidence measures from SCYM and CHC-CMIP6, this post-hoc experiment
cannot account for input data uncertainty which is likely more
substantial.

\subsection{Dual yield and risk
increases}\label{dual-yield-and-risk-increases}

Without yield exclusion, a year with claims for a risk unit would
generally decrease the subsequent \(y_{expected}\) for that risk unit.
Therefore, one may expect generally few neighborhoods and counties to
see both increased average yields and increased probability of claims
when both are calculated over a multi-year period. However, the skew for
the \emph{multi-year distributions} of yield deltas (as opposed to any
single set of annual yield deltas) grows over SSP245 as reflected
visually in our interactive tools: 2030 looks more like a normal
distribution than 2050.

\begin{longtable}[]{@{}
  >{\raggedright\arraybackslash}p{(\columnwidth - 6\tabcolsep) * \real{0.2254}}
  >{\raggedright\arraybackslash}p{(\columnwidth - 6\tabcolsep) * \real{0.2254}}
  >{\raggedright\arraybackslash}p{(\columnwidth - 6\tabcolsep) * \real{0.2958}}
  >{\raggedright\arraybackslash}p{(\columnwidth - 6\tabcolsep) * \real{0.2535}}@{}}
\caption{Frequency with which average yield and probability of claim
both increase. Counterfactual refers to simulations assuming that recent
growing conditions persist into the future. In other words, the
counterfactual assumes no further warming.
\label{tbl:dualincrease}}\label{tbl:dualincrease}\tabularnewline
\toprule\noalign{}
\begin{minipage}[b]{\linewidth}\raggedright
\textbf{Series}
\end{minipage} & \begin{minipage}[b]{\linewidth}\raggedright
\textbf{Condition}
\end{minipage} & \begin{minipage}[b]{\linewidth}\raggedright
\textbf{Neighborhoods}
\end{minipage} & \begin{minipage}[b]{\linewidth}\raggedright
\textbf{Counties}
\end{minipage} \\
\midrule\noalign{}
\endfirsthead
\toprule\noalign{}
\begin{minipage}[b]{\linewidth}\raggedright
\textbf{Series}
\end{minipage} & \begin{minipage}[b]{\linewidth}\raggedright
\textbf{Condition}
\end{minipage} & \begin{minipage}[b]{\linewidth}\raggedright
\textbf{Neighborhoods}
\end{minipage} & \begin{minipage}[b]{\linewidth}\raggedright
\textbf{Counties}
\end{minipage} \\
\midrule\noalign{}
\endhead
\bottomrule\noalign{}
\endlastfoot
2030 & Counterfactual & 3.6\% & 2.0\% \\
2050 & Counterfactual & 3.7\% & 1.9\% \\
2030 & SSP245 & 1.5\% & 1.5\% \\
2050 & SSP245 & 12.7\% & 9.8\% \\
\end{longtable}

All that in mind, Table \ref{tbl:dualincrease} shows that our
simulations report 13\% of neighborhoods and 10\% of counties seeing
both increased average yields and increased claims rates together when
calculated across the entire SSP245 2050 series\footnote{We use geohash
  center to determine county (FCC 2024). To avoid noise, we consider
  increases in average yield and increases in claims rates of less than
  2\% as essentially unchanged for this specific post-hoc experiment.
  However, the gap persists between 2050 SSP245 and 2050 counterfactual
  frequencies even if this 2\% noise filter is removed.}. This likely
reflects increased year to year volatility.

\section{Expanded definitions}\label{expanded-definitions}

We next further expand our mathematical definitions from the main text.
First, covered loss is defined as actual yields dropping below coverage
level.

\begin{equation}l = max(c * y_{expected} - y_{actual}, 0)\label{eq:loss1}\end{equation}

This can be described as a percentage of that covered yield within some
contexts where helpful.

\begin{equation}l_{\%} = max(\frac{y_{expected} - y_{actual}}{y_{expected}} - c, 0)\label{eq:loss2}\end{equation}

Furthermore, note that \(y_{expected}\) is technically defined as the
last ten years of yield for a crop. However, in practice, this may not
be calendar years due to factors like crop rotations or due to farms
with insufficient yield history.

\begin{equation}y_{expected} = \frac{y_{historic}[-d:]}{d}\label{eq:expected1}\end{equation}
\begin{equation}y_{expected} = \frac{y_{historic}[-min(10, |y_{historic}|):]}{min(10, |y_{historic}|)}\label{eq:expected2}\end{equation}

Next, the probability of experiencing a loss that may incur a Yield
Protection claim (\(p_{l}\)) may be defined a few different ways
depending on data available at the potin in the pipeline.

\begin{equation}p_{l} = P(l > 0) = P(c * y_{expected} - y_{actual} > 0)\label{eq:ploss1}\end{equation}
\begin{equation}p_{l} = P(\frac{y_{actual} - y_{expected}}{y_{expected}} < c - 1)\label{eq:ploss2}\end{equation}
\begin{equation}p_{l} = P(y_{\Delta\%} < c - 1)\label{eq:ploss3}\end{equation}

Finally, the severity (\(s\)) of a loss may also take multiple forms.

\begin{equation}s = \frac{l}{y_{expected}}\label{eq:severity1}\end{equation}
\begin{equation}s = \max(c - \frac{y_{actual}}{y_{expected}}, 0)\label{eq:severity2}\end{equation}
\begin{equation}s = \max(-1 * y_{\Delta\%} - (1 - c), 0)\label{eq:severity3}\end{equation}

Our interactive tools further explain these formulations and how they
fit together to define preimums and claims.

\section{Interactive tools}\label{interactive-tools}

Finally, we further describe our interactive tools. In crafting these
``explorable explanations'' (Victor 2011) listed in Table
\ref{tbl:apps}, we draw analogies to micro-apps (Bridgwater 2015) or
mini-games (DellaFave 2014) in which the user encounters a series of
small experiences that, each with distinct interaction and objectives,
can only provide minimal instruction (Brown 2024). As these very brief
visualization experiences cannot take advantage of design techniques
like Hayashida-style tutorials (A. S. Pottinger and Zarpellon 2023),
they rely on simple ``loops'' (Brazie 2024) for immediate
``juxtaposition gratification'' (JG) (JM8 2024), showing fast
progression after minimal input.

\begin{longtable}[]{@{}
  >{\raggedright\arraybackslash}p{(\columnwidth - 6\tabcolsep) * \real{0.2117}}
  >{\raggedright\arraybackslash}p{(\columnwidth - 6\tabcolsep) * \real{0.2423}}
  >{\raggedright\arraybackslash}p{(\columnwidth - 6\tabcolsep) * \real{0.2607}}
  >{\raggedright\arraybackslash}p{(\columnwidth - 6\tabcolsep) * \real{0.2853}}@{}}
\caption{Overview of explorable explanations.
\label{tbl:apps}}\label{tbl:apps}\tabularnewline
\toprule\noalign{}
\begin{minipage}[b]{\linewidth}\raggedright
\textbf{Simulator}
\end{minipage} & \begin{minipage}[b]{\linewidth}\raggedright
\textbf{Question}
\end{minipage} & \begin{minipage}[b]{\linewidth}\raggedright
\textbf{Loop}
\end{minipage} & \begin{minipage}[b]{\linewidth}\raggedright
\textbf{JG}
\end{minipage} \\
\midrule\noalign{}
\endfirsthead
\toprule\noalign{}
\begin{minipage}[b]{\linewidth}\raggedright
\textbf{Simulator}
\end{minipage} & \begin{minipage}[b]{\linewidth}\raggedright
\textbf{Question}
\end{minipage} & \begin{minipage}[b]{\linewidth}\raggedright
\textbf{Loop}
\end{minipage} & \begin{minipage}[b]{\linewidth}\raggedright
\textbf{JG}
\end{minipage} \\
\midrule\noalign{}
\endhead
\bottomrule\noalign{}
\endlastfoot
Rates & What factors influence the price and subsidy of a policy? &
Iteratively change variables to increase subsidy. & Improving on
previous hypotheses. \\
Hyper-Parameter & How do hyper-parameters impact regressor performance?
& Iteratively change neural network hyper-parameters to see influence on
validation set performance. & Improving on previous hyper-parameter
hypotheses. \\
Distributional & How do overall simulation results change under
different simulation parameters? & Iterative manipulation of parameters
(geohash size, event threshold, year) to change loss probability and
severity. & Deviating from the study's main results. \\
Neighborhood & How do simulation results change across geography and
climate conditions? & Inner loop changing simulation parameters to see
changes in neighborhood outcomes. Outer loop of observing changes across
different views. & Identifying neighborhood clusters of concern. \\
Claims & How do different regulatory choices influence grower behavior?
& Iteratively change production history to see which years result in
claims under different regulatory schemes. & Redefining policy to
improve yield stability. \\
\end{longtable}

Following Unwin (2020), our custom tools first serve as internal
exploratory graphics enabling the insights detailed in our results
before acting as a medium for sharing our work.

\subsection{Internal use}\label{internal-use}

First, these tools were built during our own internal exploration of
data with Table \ref{tbl:insights} outlining specific observations we
attribute to our use of these tools.

\begin{longtable}[]{@{}
  >{\raggedright\arraybackslash}p{(\columnwidth - 2\tabcolsep) * \real{0.2759}}
  >{\raggedright\arraybackslash}p{(\columnwidth - 2\tabcolsep) * \real{0.7241}}@{}}
\caption{Observations we made from our own tools in the ``exploratory''
graphic context of Unwin (2020).
\label{tbl:insights}}\label{tbl:insights}\tabularnewline
\toprule\noalign{}
\begin{minipage}[b]{\linewidth}\raggedright
\textbf{Simulator}
\end{minipage} & \begin{minipage}[b]{\linewidth}\raggedright
\textbf{Observation}
\end{minipage} \\
\midrule\noalign{}
\endfirsthead
\toprule\noalign{}
\begin{minipage}[b]{\linewidth}\raggedright
\textbf{Simulator}
\end{minipage} & \begin{minipage}[b]{\linewidth}\raggedright
\textbf{Observation}
\end{minipage} \\
\midrule\noalign{}
\endhead
\bottomrule\noalign{}
\endlastfoot
Distributional & Dichotomy of changes to yield and changes to loss
risk. \\
Claims & Issues of using average for \(y_{expected}\) (FCIC 2020). \\
Neighborhood & Geographic bias of impact and model output relationships
with broader climate factors. \\
Hyper-parameter & Model resilience to removing individual inputs. \\
\end{longtable}

Altogether, these tools serve to support our exploration of our modeling
such as different loss thresholds for other insurance products, finding
relationships of outcomes to different climate variables, understanding
interaction with insurance mechanisms, answering geographically specific
questions, and modification of machine learning parameters to understand
performance.

\subsection{Workshops}\label{workshops}

In addition to supporting our finding of our own conclusions, we release
this software publicly at https://ag-adaptation-study.pub/. For example,
possible use of these tools may include workshop activity. To support
use of these tools as supplement to this paper, we made the following
changes\footnote{These were implemented in response to our work's
  participation in a 9 person ``real-world'' workshop session
  encompassing scientists and engineers which was intended to improve
  these tools specifically through active co-exploration limited to
  these study results. We collect information about the tool only and
  not generalizable knowledge about users or these patterns, falling
  under ``quality assurance'' activity. IRB questionnaire on file. This
  was \emph{not} a public workshop or a formalized academic conference
  presentation.}:

\begin{itemize}
\tightlist
\item
  We elect to alternate between presentation and interaction similar to
  A. S. Pottinger et al. (2023). However, we added the rates simulator
  to further improve presentation of the rate setting process due to the
  complexities of crop insurance, dynamics previously explained in
  static diagrams.
\item
  Our single loop (Brazie 2024) designs may be better suited to the
  limited timeframe of a workshop. Therefore, we now let facilitators
  hold the longer two loop neighborhood simulator till the end by
  default.
\item
  While the JG design (JM8 2024) expects discussion to contrast
  different results sets and configurations of models, the
  meta-parameter visualization specifically relies heavily on memory so
  we now offer a ``sweep'' button for facilitators to show all results
  at once.
\end{itemize}

Later work may more broadly explore this design space through controlled
experimentation (Lewis 1982) or diary studies (Shneiderman and Plaisant
2006).

\section*{Works cited}\label{works-cited}
\addcontentsline{toc}{section}{Works cited}

\phantomsection\label{refs}